\def\arxivversion{2}
\documentclass[square]{ws-procs11x85}
\usepackage{ws-rotating}
\AtBeginDvi{}
\usepackage{ws-procs-thm}
\usepackage{tikz}
\usepackage{pgf-pie}
\usepackage{fontspec}
\usepackage{xeCJK}
\usepackage{twemojis}
\usepackage{xcolor}
\usepackage{array}
\usepackage{listings}
\usepackage[most]{tcolorbox}
\usetikzlibrary{arrows.meta,fit,positioning,shapes.geometric}

\ifdefined\arxivversion
\else
\newfontfamily\russianfont[
  Script=Cyrillic,
  Language=Russian
]{Times New Roman}
\newCJKfontfamily\chinesefont{Microsoft YaHei}
\newCJKfontfamily\japanesefont{Yu Gothic}
\newCJKfontfamily\koreanfont{Malgun Gothic}

\newfontfamily\laofont{Leelawadee UI}
\newfontfamily\thaifont{Leelawadee UI}

\newfontfamily\burmesefont{Myanmar Text}

\newfontfamily\tamilfont{Nirmala UI}
\fi

\definecolor{MeddiesPromptFrame}{RGB}{33, 112, 120}
\definecolor{MeddiesPromptBack}{RGB}{248, 252, 252}
\definecolor{MeddiesPromptTitle}{RGB}{227, 242, 243}
\definecolor{MeddiesChartPrimary}{RGB}{86, 180, 233}
\definecolor{MeddiesChartSecondary}{RGB}{230, 159, 0}
\definecolor{MeddiesChartNeutral}{RGB}{0, 158, 115}

\lstdefinestyle{meddiesprompt}{
  basicstyle=\ttfamily\scriptsize,
  breaklines=true,
  breakatwhitespace=false,
  breakindent=0pt,
  breakautoindent=false,
  columns=fullflexible,
  keepspaces=true,
  showstringspaces=false
}

\makeatletter
\newcommand{\promptfigurecaption}[2]{%
  \refstepcounter{figure}%
  \par\smallskip\noindent{\small\textbf{Fig.~\thefigure.} #1}\label{#2}\par\medskip%
}
\makeatother

\begin{document}

\ifdefined\arxivversion
\else
\begingroup
\thispagestyle{empty}
\noindent August 3, 2026

\vspace{2em}
\noindent PSB 2027 Program Committee

\vspace{2em}
\noindent Dear Program Committee,

\vspace{1em}
\noindent We submit the manuscript ``Meddies-PII: A Multilingual Framework for Personally Identifiable Information Extraction in Clinical De-identification'' for consideration at the Pacific Symposium on Biocomputing 2027.

\vspace{1em}
\noindent We request that the paper be reviewed for the session ``NLP Methods for Embedding Real-World Clinical Knowledge in LLMs: Towards Responsible, Transformative Medical AI.''

\vspace{1em}
\noindent The corresponding author is Huy Hoang Ha (\texttt{hoangha@meddies.ai}).

\vspace{1em}
\noindent We confirm that this submission contains original, unpublished results and is not currently under consideration elsewhere. All co-authors concur with the contents of the paper and its submission to PSB 2027.

\vspace{1em}
\noindent Large language models were used to assist with language editing and manuscript revision. The authors reviewed the resulting text and take full responsibility for the paper's content.

\vspace{2em}
\noindent Sincerely,

\vspace{1em}
\noindent Le Linh Uyen*, Ho Dinh Tri*, Christian Hoang, and Huy Hoang Ha\\
Meddies AI, Vietnam
\vfill
\clearpage
\endgroup
\fi

\title{Meddies-PII: A Multilingual Framework for Personally Identifiable Information Extraction in Clinical De-identification}

\author{
Le Linh Uyen$^{1,*}$,
Ho Dinh Tri$^{1,*}$,
Christian Hoang$^{1}$,
and Huy Hoang Ha$^{1,\dag}$
}

\address{
$^{1}$Meddies AI, Vietnam\\
$^{*}$These authors contributed equally.\\
$^{\dag}$Corresponding author: hoangha@meddies.ai
}

\begin{abstract}
Clinical de-identification depends on accurately identifying personally identifiable information (PII). However, manually annotated datasets are costly to build, and existing synthetic alternatives often provide limited generation details or use relatively simple synthesis strategies. We introduce Meddies-PII-Dataset, which contains one million synthetic clinical documents across seventeen languages and nine PII labels. Attribute-conditioned prompts generate the documents, and thirteen deterministic gates check their structural and annotation validity. To assess the dataset's utility, we train Meddies-PII-Model, a BIOES token classifier. We compare it with existing PII extraction systems using exact-match entity-level F1. Meddies-PII-Model achieves the highest score among the evaluated systems on all reported benchmarks, with a mean F1 of 0.827 across fifteen external benchmarks compared with 0.658 for the strongest baseline. Upon acceptance, we will publicly release the dataset, benchmark, model, generation framework, and evaluation code to support research on multilingual clinical de-identification.
\end{abstract}


\ifdefined\arxivversion
\copyrightinfo{Preprint of an article submitted for consideration in Pacific Symposium on Biocomputing \copyright\ 2026 World Scientific Publishing Co., Singapore, \url{http://psb.stanford.edu/}.}
\else
\copyrightinfo{\copyright\ 2026 The Authors. Open Access chapter published by World Scientific Publishing Company and distributed under the terms of the Creative Commons Attribution Non-Commercial (CC BY-NC) 4.0 License.}
\fi

\section{Introduction}

Named Entity Recognition (NER) is a fundamental task in natural language processing. In privacy-sensitive domains such as healthcare, however, the primary objective is identifying personally identifiable information (PII), a specialized subset of named entities whose disclosure may compromise patient privacy \cite{savkin-etal-2025-spy}. Although PII extraction is commonly formulated as an NER task, it differs from general-purpose NER in both its ontology and application objectives. Consequently, data-driven PII systems require dedicated datasets rather than relying solely on conventional NER corpora. Existing PII datasets are predominantly manually annotated, making them expensive to construct and limiting their scale, language coverage, and domain diversity \cite{neamatullah2008automated,uzuner2007evaluating,stubbs2015automated}. Although several synthetic PII datasets have recently been proposed, most either do not disclose their generation pipeline or rely on relatively simple prompting strategies \cite{ai4privacy_openpii_1_5m_2026,gretel-pii-docs-en-v1,nemotron-pii}. 

We propose a multilingual synthetic data generation framework that combines attribute-conditioned prompting with deterministic validation to construct Meddies-PII-Dataset. The framework controls language, document type, text format, scenario, ontology constraints, and challenging edge cases to generate diverse and structurally complex documents. Generated samples that violate the declared requirements are repaired or rejected before inclusion in the dataset. We further train a token-classification model to evaluate the utility of the resulting data. We evaluate the model on fifteen external benchmarks and the Meddies-PII Benchmark. Our benchmark combines standard and challenging examples under the Meddies-PII ontology to assess performance across different levels of difficulty.

This paper makes three contributions:
\begin{arabiclist}[(3)]
\item We introduce \textbf{Meddies-PII-Dataset}, a synthetic clinical PII dataset containing one million documents across seventeen languages and annotated with a nine-label PII ontology.
\item We present a controlled multilingual synthesis framework that combines multiple document and generation attributes with deterministic structural and annotation validation.
\item We introduce \textbf{Meddies-PII-Model}, a BIOES token classifier, and show that it achieves the highest performance on all evaluated external benchmarks and our benchmark.
\end{arabiclist}

\section{Related Works}

\subsection{PII Datasets}

\paragraph{Manually annotated datasets} Manually annotated PII datasets support supervised training and evaluation, but expert annotation makes them costly to construct and difficult to scale \cite{deleger2014preparing}. Early clinical de-identification resources focused primarily on English medical records and supported influential shared tasks and system evaluations \cite{neamatullah2008automated,uzuner2007evaluating,stubbs2015automated}. Non-English clinical resources include corpora for French clinical notes and the German GRASCCO corpus \cite{grouin2014identification,modersohn2022grascco}. Sensitive-entity datasets have also been developed for Spanish legal text, showing progress beyond the clinical domain \cite{de-gibert-bonet-etal-2022-spanish}. More recent multilingual resources broaden this coverage, but manually annotated datasets still span relatively few languages and domains \cite{baroud2026multigrascco}.

\paragraph{Synthetic datasets} Synthetic generation provides a scalable alternative to manually annotated datasets, and several large PII datasets have recently been released \cite{ai4privacy_openpii_1_5m_2026,gretel-pii-docs-en-v1,nemotron-pii}. However, their generation procedures are not always described in enough detail to reproduce the controls used for document diversity, label coverage, and difficult examples. SPY \cite{savkin-etal-2025-spy} documents its workflow more clearly, including iterative placeholder insertion, synthetic entity swapping, and the addition of unrelated entities. Meddies-PII-Dataset addresses this reproducibility gap through an independently designed multilingual pipeline that explicitly controls document and generation attributes and applies deterministic validation before accepting samples.

\subsection{PII Models}

\paragraph{NER Techniques}

NER systems identify text spans and assign each span an entity label. Many neural approaches formulate this task as sequence labeling, predicting a label for each token and optionally using a Conditional Random Field (CRF) to model dependencies between adjacent labels \cite{souza2019portuguese}. Transformer encoders such as BERT improve these predictions by representing each token in its surrounding context \cite{devlin2019bert}. Span-based methods instead score candidate spans directly, allowing the model to predict entity boundaries and labels together \cite{eberts2019span}. More recent generalist models, including GLiNER and GLiNER2, use schema-driven interfaces to recognize entity types beyond a single fixed taxonomy \cite{zaratiana2023gliner,zaratiana2025gliner2efficientmultitaskinformation}.

\paragraph{PII Techniques} PII extraction specializes NER by mapping detected spans to privacy-related labels used for de-identification. Generalist systems such as GLiNER2 can treat these labels as a requested extraction schema, while dedicated systems such as OpenMed-PII and OpenAI Privacy Filter are designed specifically for PII detection \cite{zaratiana2025gliner2efficientmultitaskinformation,openmed-pii-2026,openai2026privacyfilter}. OpenAI Privacy Filter uses bidirectional token classification and span decoding over a fixed taxonomy of privacy labels \cite{openai2026privacyfilter}. This formulation combines contextual token representations with explicit entity-boundary prediction. Following this approach, we train Meddies-PII-Model on Meddies-PII-Dataset using BIOES tagging and our nine-label PII ontology.

\section{Meddies-PII-Dataset}

\subsection{Overview}

\begin{figure}[!htbp]
\centering
\includegraphics[width=1\textwidth]{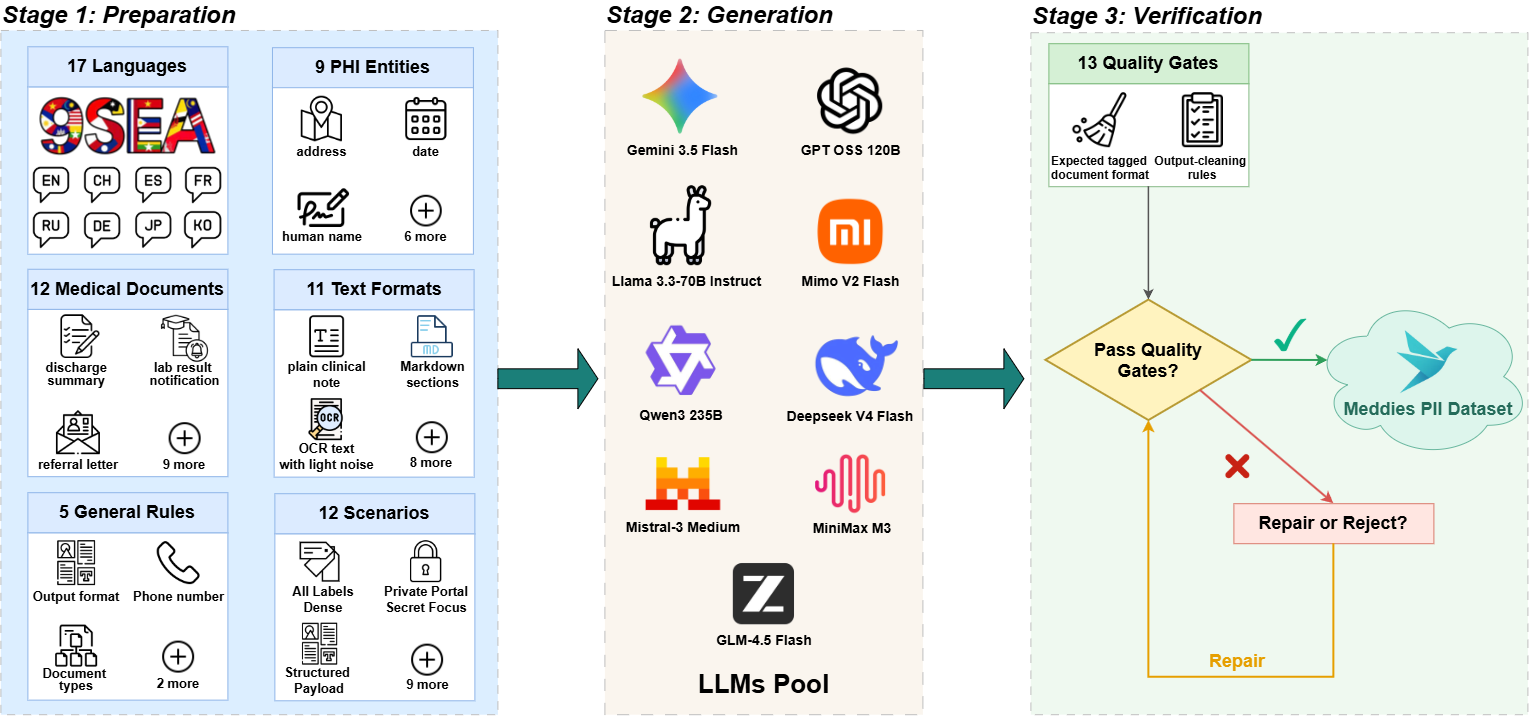}
\caption{Synthetic data generation and verification pipeline to create \textbf{Meddies-PII-Dataset}.}
\label{fig:data-generation-pipeline}
\vspace{1em}
\end{figure}

Manually annotated PII datasets provide expert-reviewed labels for training and evaluation, but their cost limits their scale, language coverage, and domain diversity \cite{deleger2014preparing,neamatullah2008automated}. Synthetic generation offers greater scalability, although insufficiently documented controls and validation procedures can make existing datasets difficult to reproduce \cite{ai4privacy_openpii_1_5m_2026,gretel-pii-docs-en-v1,nemotron-pii}. Meddies-PII-Dataset addresses this trade-off through a multilingual generation framework that makes its generation attributes and deterministic validation process explicit. Figure \ref{fig:data-generation-pipeline} provides an overview of the complete pipeline.

\paragraph{Dataset composition}

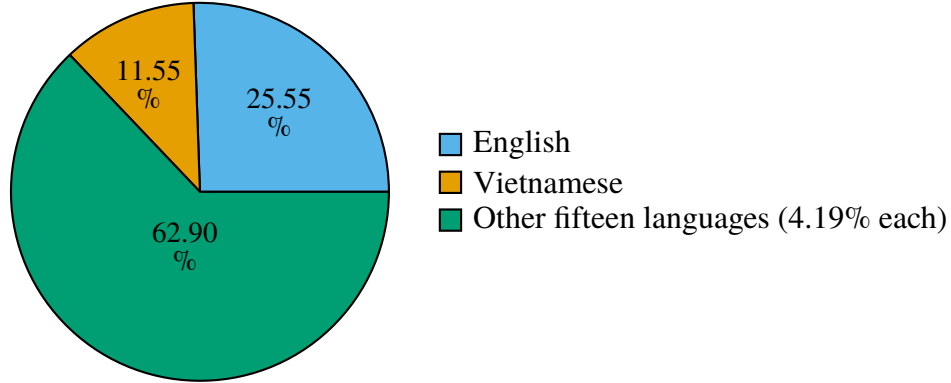
\begin{figure}[!htbp]
\centering
\hspace*{0.30in}
\begin{tikzpicture}
\pie[
    sum=100,
    text=legend,
    hide number=true,
    radius=2.50,
    color={MeddiesChartPrimary,MeddiesChartSecondary,MeddiesChartNeutral}
]{25.55/English,11.55/Vietnamese,62.90/Other fifteen languages (4.19\% each)}
\node at (1.05,1.05) {\shortstack{\small 25.55 \\ \small\%}};
\node at (-0.68,1.43) {\shortstack{\small 11.55 \\ \small\%}};
\node at (-0.20,-0.72) {\shortstack{\small 62.90 \\ \small\%}};
\end{tikzpicture}
\caption{Language distribution of Meddies-PII-Dataset. The other fifteen languages each contain 41,934 samples (4.19\%).}
\label{fig:language-distribution}
\end{figure}

Meddies-PII-Dataset contains one million synthetic documents across seventeen languages: English, Vietnamese, Chinese, Japanese, Lao, Tamil, Thai, Burmese, Filipino, Russian, Spanish, Malay, Indonesian, Portuguese, French, Korean, and German. After generating the initial collection, we reviewed its samples and identified opportunities to improve data quality. We then refined the generation process and created a second collection of higher-quality samples. The final dataset mixes selected samples from both collections to reach one million documents while preserving the target language distribution and broader coverage. This design is consistent with sample-wise pretraining data mixing research that jointly considers quality and diversity when determining the resulting data distribution \cite{xi-etal-2025-samplemix}. The language selection emphasizes Southeast Asian languages, several of which have limited representation in existing PII datasets. Because widely spoken international languages are also used across Southeast Asian countries, we extend the dataset with prominent languages beyond the region, yielding coverage across different scripts, language families, and resource levels. Figure \ref{fig:language-distribution} summarizes the language distribution.

\paragraph{PII labels}

We define a nine-label PII ontology informed by the identifiers and distinguishing attributes described in ISO/IEC 29100:2024 \cite{iso29100-2024}:
\begin{itemize}
    \item \texttt{address}: street addresses, postal codes, coordinates, and care locations;
    \item \texttt{company\_name}: hospitals, clinics, departments, companies, and other organizations;
    \item \texttt{date}: birth dates, admission and discharge dates, and other specific datetimes linked to a person or encounter;
    \item \texttt{email\_address}: email addresses;
    \item \texttt{human\_name}: patient, clinician, and other person names;
    \item \texttt{id\_number}: medical record numbers, government identifiers, passport and licence numbers, IP addresses, and account numbers;
    \item \texttt{phone\_number}: telephone and fax numbers;
    \item \texttt{private\_url}: access-bearing patient-portal, result, signed, or private-record URLs;
    \item \texttt{secret}: passwords, API keys, authentication or session tokens, cookies, personal identification numbers, and one-time passwords.
\end{itemize}

\paragraph{Meddies-PII Benchmark} Our benchmark contains 5,100 validated documents distributed evenly across the seventeen languages, with 300 documents per language. For each language, 100 documents assess PII extraction under standard conditions, while 200 emphasize adversarial surface forms. Each adversarial document combines four to five of thirteen perturbation techniques. These include spoken digits and email symbols, phonetic spelling, unusual spacing, entity-internal line breaks, truncation, partial masking, OCR-like substitutions, Unicode confusables, and concatenated values. These perturbations preserve the annotated PII spans while testing robustness to noise commonly encountered in speech transcripts, scanned records, messages, and administrative documents.

\subsection{Dataset Creation}

As shown in Figure \ref{fig:data-generation-pipeline}, dataset creation proceeds through three stages: Preparation, Generation, and Verification. For each sample, the framework prepares a controlled prompt, sends it to one model from the generation pool, and validates the resulting document before accepting it into the dataset.

\paragraph{Stage 1: Preparation} Each generation prompt combines six attribute groups that control the content and structure of a sample. Medical documents determine the workflow and content, while text formats determine how that content is expressed. General rules maintain annotation consistency, and scenarios control the operational context and difficulty. Together, these groups capture complementary sources of variation in clinical PII data. Randomly combining attribute values produces varied prompts at scale, an approach supported by evidence that attributed prompts improve synthetic-data diversity and downstream performance while reducing query cost \cite{yu2023large}. We refined the final set of options through project discussions to balance realistic clinical and administrative use cases with broad label and structural coverage. Figure \ref{fig:meddies_user_prompt} shows the prompt template.

\begin{enumerate}
    \item \textbf{Languages:} Specifies the target language of the generated document.
    \item \textbf{PII labels:} Nine labels specify the PII categories and required label coverage for each document.
    \item \textbf{Medical documents:} Represents PII in common clinical and administrative workflows. The twelve documents are discharge summary; outpatient note; lab-result notification; patient portal access log; insurance claim note; referral letter; pharmacy refill request; appointment reminder; telehealth triage note; medical bill or invoice; FHIR-like Patient and Encounter resource; and HL7-style admission or update message.
    \item \textbf{Text formats:} Captures how equivalent information changes across prose, structured records, logs, transcripts, and noisy text. The eleven formats are plain clinical note; Markdown sections; messy nurse or administrative note; table-like rows; JSON-like object; FHIR-like JSON; HL7-like pipe-delimited text; patient portal audit log; voice-scribe transcript; lightly degraded OCR text; and clinical dialogue transcript.
    \item \textbf{General rules:} Maintains consistent generation and annotation across combinations of attributes. The five rule groups govern the output format, phone-number construction, private URLs, secrets, and compatibility between medical documents and text formats.
    \item \textbf{Scenarios:} Places PII within realistic operational contexts while controlling label density and difficulty. The twelve scenarios are All Labels Dense; Private Portal Secret Focus; Adversarial Obfuscation; Public Negative Contrast; Structured Payload; General Consumer Admin Support; HR, Employer, School, Insurance, or Telecom Admin; Code, Log, Security, or Account Recovery; Adversarial Formatting; Structured General Payload; One-Hop Deferred Clue; and Public Negative URL and Data Contrast.
\end{enumerate}

\begin{tcolorbox}[
    enhanced,
    colback=MeddiesPromptBack,
    colframe=MeddiesPromptFrame,
    sharp corners,
    boxrule=0.9pt
]
\begin{lstlisting}[style=meddiesprompt, language=Python]
user_prompt = f"""
Generate one synthetic {language} {document_type} in {text_format} format.

Scenario: {scenario_name}
{scenario_description}

Hard constraints:
1. Use at least {min_spans} tagged PII spans.
2. Use at least {min_unique_labels} unique Meddies labels.
3. Required labels for this sample: {required_labels}.
etc.

Required-label coverage before you answer: {", ".join(f"{label}=present" for label in resolved_required_labels) or "none"}.
Minimum coverage before you answer: at least {min_spans} bracketed tags and at least {min_unique_labels} unique labels.

Remember: label syntax must be `[value]<label>` and labels must be one of: {", ".join(f"<{label}>" for label in PII_LABELS)}.
"""
\end{lstlisting}
\end{tcolorbox}

\promptfigurecaption
    {User-prompt template for Meddies-PII-Dataset generation. Braced fields are populated from the sampled generation attributes and validation requirements.}
    {fig:meddies_user_prompt}

\paragraph{Hard constraints} The hard-constraint block converts the sampled attributes into explicit generation requirements. It specifies the minimum number of tagged spans and unique labels, lists labels that must appear, and controls when private URLs, secrets, negative examples, and adversarial perturbations are required. It also enforces the exact \texttt{[value]<label>} syntax, complete entity boundaries, document-only output, and a final coverage self-check. These constraints guide the model toward drafts that can be checked by the deterministic gates in Stage 3, although a draft is accepted only after passing verification.

\paragraph{Stage 2: Generation} The generation pool contains nine large language models (LLMs) from different model families. These are Gemini 3.5 Flash \cite{googledeepmind2026gemini35flash}, GLM-4.5-Flash \cite{zeng2025glm}, Llama 3.3 70B Instruct \cite{meta2024llama33}, MiMo-V2-Flash \cite{xiao2026mimo}, Mistral Medium 3 \cite{mistralai2025mistralmedium3}, gpt-oss-120b \cite{agarwal2025gpt}, DeepSeek-V4-Flash \cite{deepseekai2026v4flash}, Qwen3-235B-A22B \cite{yang2025qwen3}, and MiniMax M3 \cite{minimax2026m3}. For each sample, the framework sends the assembled prompt to one selected LLM from this pool. Distributing generation across model families reduces the pipeline's dependence on the behavior of a single generator.

\paragraph{Stage 3: Verification} Each generated draft enters deterministic verification. Verification uses regular expressions, a deterministic character-level parser, rule-based checks, and SHA-256 hashing. Regular expressions check annotation syntax and language-specific Unicode ranges. The parser reconstructs the raw document and verifies span offsets, while rule-based checks validate phone numbers. SHA-256 hashes detect duplicate documents. Together, thirteen gates check label coverage, span count, document length, offsets, phone-number structure, native-script presence, and duplication. Drafts that pass are accepted, while failed drafts enter the repair workflow. API failures and empty content are rejected because they contain no recoverable document, while unknown scenarios are rejected because the metadata needed to interpret and validate the document is unavailable. For the remaining drafts, conservative regex repair fixes only explicit labels with unambiguous value boundaries, such as misplaced brackets, spacing around tags, or unwrapped structured-field values. Repaired drafts are then revalidated and accepted only if they pass every gate and are unique. Drafts with missing or unknown labels, too few spans, invalid lengths or offsets, templated phone numbers, or missing native script are rejected because correcting them would require inventing content, labels.

\section{Evaluation}

\subsection{Evaluation Settings}

\paragraph{Evaluation Protocol} We compare Meddies-PII-Model with four publicly available PII extraction systems: OpenAI Privacy Filter \cite{openai2026privacyfilter}, GLiNER2 \cite{zaratiana2025gliner2efficientmultitaskinformation}, SuperClinical-Large-434M-v1 \cite{openmed-pii-2026}, and LFM2.5-Encoder-350M-PII-Detector \cite{liquidai2026lfm25pii}. The external evaluation includes twelve language subsets from OpenPII 1.5M \cite{ai4privacy_openpii_1_5m_2026}, together with CredData \cite{sr-cred21}, Gretel PII Masking \cite{gretel-pii-docs-en-v1}, and Nemotron-PII \cite{nemotron-pii}. We additionally evaluate every system on the Meddies-PII Benchmark. Labels produced by each system are mapped to the shared PII ontology before scoring. We report exact-match entity-level micro-F1 by pooling entity counts across documents. A predicted entity is a true positive only when both its PII label and character boundaries exactly match a reference entity. Predictions with an incorrect label or boundary are counted as false positives, while unmatched reference entities are counted as false negatives. Using the pooled counts, the metric is computed as
\begin{equation}
P=\frac{TP}{TP+FP},\qquad
R=\frac{TP}{TP+FN},\qquad
F_1=\frac{2PR}{P+R}=\frac{2TP}{2TP+FP+FN}.
\end{equation}

\paragraph{Model Development} Our initial model development treated PII extraction as autoregressive structured generation. We first applied supervised fine-tuning (SFT) and subsequently reinforcement learning (RL), but sequential decoding made inference slow and the resulting extraction quality was not competitive. We therefore adopted the one-pass token-classification approach used by OpenAI Privacy Filter \cite{openai2026privacyfilter}. This approach classifies each token using bidirectional document context instead of generating entities token by token. BIOES marks tokens as Beginning, Inside, End, Single-token, or Outside (O) relative to an entity. Meddies-PII-Model fine-tunes LFM2.5-350M with a classifier containing 37 output classes: O and four states for each of the nine PII labels. A linear classification head predicts a BIOES state for each token. Constrained Viterbi decoding then converts the predicted sequence into character-level entity spans while enforcing valid BIOES transitions \cite{lester-etal-2020-constrained}.

\subsection{Results}

\begin{table*}[!htbp]
\tbl{Exact-match entity-level micro-F1 scores. The best result for each benchmark is shown in bold.}
{\resizebox{\textwidth}{!}{
\begin{tabular}{l*{5}{>{\centering\arraybackslash}m{2.45cm}}}
\toprule
\textbf{Benchmark} &
\textbf{\shortstack{Meddies PII\\Model}} &
\textbf{\shortstack{OpenAI Privacy\\Filter}} &
\textbf{GLiNER2} &
\textbf{\shortstack{SuperClinical\\Large 434M}} &
\textbf{\shortstack{LFM2.5 Encoder\\350M PII Detector}} \\
\colrule

\multicolumn{6}{l}{\textit{External evaluation}} \\
OpenPII 1.5M (de)  & \textbf{0.827} & 0.716 & 0.663 & 0.620 & 0.610 \\
OpenPII 1.5M (en)  & \textbf{0.878} & 0.745 & 0.687 & 0.656 & 0.568 \\
OpenPII 1.5M (es)  & \textbf{0.870} & 0.794 & 0.732 & 0.683 & 0.655 \\
OpenPII 1.5M (fil) & \textbf{0.848} & 0.667 & 0.690 & 0.584 & 0.600 \\
OpenPII 1.5M (fr)  & \textbf{0.844} & 0.738 & 0.674 & 0.604 & 0.611 \\
OpenPII 1.5M (id)  & \textbf{0.848} & 0.664 & 0.688 & 0.566 & 0.627 \\
OpenPII 1.5M (ja)  & \textbf{0.750} & 0.549 & 0.366 & 0.493 & 0.479 \\
OpenPII 1.5M (ko)  & \textbf{0.739} & 0.554 & 0.322 & 0.545 & 0.460 \\
OpenPII 1.5M (ms)  & \textbf{0.839} & 0.640 & 0.692 & 0.547 & 0.587 \\
OpenPII 1.5M (pt)  & \textbf{0.855} & 0.756 & 0.692 & 0.687 & 0.651 \\
OpenPII 1.5M (vi)  & \textbf{0.880} & 0.634 & 0.628 & 0.499 & 0.615 \\
OpenPII 1.5M (zh)  & \textbf{0.816} & 0.572 & 0.453 & 0.428 & 0.548 \\
CredData (en)      & \textbf{0.590} & 0.370 & 0.357 & 0.150 & 0.263 \\
Gretel (en)        & \textbf{0.872} & 0.697 & 0.588 & 0.715 & 0.729 \\
Nemotron (en)      & \textbf{0.956} & 0.778 & 0.692 & 0.878 & 0.574 \\
\textbf{External mean} & \textbf{0.827} & 0.658 & 0.595 & 0.577 & 0.572 \\

\colrule
\textbf{Meddies-PII Benchmark} & \textbf{0.878} & 0.540 & 0.476 & 0.244 & 0.269 \\

\colrule
\textbf{Overall mean} &
\textbf{0.833} &
0.644 &
0.581 &
0.538 &
0.536 \\
\botrule
\end{tabular}
}}
\label{tab:benchmark-results}
\end{table*}

\paragraph{Overall comparison} Table \ref{tab:benchmark-results} reports exact-match entity-level micro-F1 for the evaluated systems. Meddies-PII-Model ranks first on every benchmark shown. Across the fifteen external benchmarks, it achieves an unweighted mean F1 of 0.827, exceeding OpenAI Privacy Filter, the strongest baseline by external mean, by 0.169. When the Meddies-PII Benchmark is included, the overall means are 0.833 and 0.644, respectively. The consistent advantage across both external and in-domain evaluations suggests that the improvement is not confined to the distribution used to develop Meddies-PII-Model. 

\paragraph{Multilingual performance} On the twelve OpenPII 1.5M language benchmarks, Meddies-PII-Model scores between 0.739 and 0.880 and ranks first in every language. Its largest margins over the strongest competing score occur for Vietnamese (0.880 versus 0.634), Chinese (0.816 versus 0.572), and Korean (0.739 versus 0.554). This pattern indicates that the external mean is not driven only by English and that the model transfers effectively across several languages and scripts. However, the lower absolute scores for Korean and Japanese show that multilingual performance is not uniform, leaving room for language-specific analysis and improvement.

\paragraph{Cross-dataset performance} Meddies-PII-Model achieves F1 scores of 0.590 on CredData, 0.872 on Gretel, and 0.956 on Nemotron. The strongest baseline scores on the same benchmarks are 0.370, 0.729, and 0.878, giving margins of 0.220, 0.143, and 0.078, respectively. Positive margins across all three datasets provide evidence of transfer beyond a single annotation source. CredData remains substantially more difficult for every system, indicating that differences in domain, document structure, or annotation policy still present a meaningful challenge.

\paragraph{Meddies-PII Benchmark} The Meddies-PII Benchmark combines standard examples with examples emphasizing adversarial surface forms. On this benchmark, Meddies-PII-Model achieves an F1 of 0.878, compared with 0.540 for the strongest baseline. The 0.338 margin shows that the model handles the ontology and perturbations represented in its target setting more effectively than the compared systems. Because this is an in-domain evaluation, we report and interpret it separately from the external mean rather than treating it as direct evidence of generalization.

Taken together, the results show consistent gains across languages and datasets, while the lower scores on CredData, Korean, and Japanese identify important directions for further evaluation.

\paragraph{Data and code availability} Upon acceptance, we will publicly release Meddies-PII-Dataset, the Meddies-PII Benchmark, Meddies-PII-Model, the generation framework, and the evaluation code.

\section{Limitations}

Meddies-PII-Dataset is entirely synthetic and may not capture all linguistic variation, documentation practices, and annotation ambiguity found in real-world clinical records. Its deterministic gates enforce the declared structural and annotation requirements, but they do not directly measure the semantic naturalness of every generated document. In addition, the present evaluation focuses on synthetic PII benchmarks and Meddies-PII Benchmark data; evaluation on appropriately governed real clinical records remains an important direction for future work.

\section{Conclusion}

We presented \textbf{Meddies-PII-Dataset}, a synthetic clinical PII dataset containing one million documents across seventeen languages and nine PII labels. After reviewing the initial generation, we refined the process, generated a second collection of higher-quality samples, and mixed both collections to preserve the target distribution and broader coverage. The improved process controls multiple document and scenario attributes and selects one generator per sample from a pool of nine models. Thirteen deterministic gates then validate each sample before acceptance. \textbf{Meddies-PII-Model} achieves the highest performance among the evaluated systems across all reported external benchmarks and the \textbf{Meddies-PII Benchmark}, demonstrating the downstream utility of the generated data. By making the dataset, benchmark, model, generation framework, and evaluation code publicly available upon acceptance, we aim to support further research on multilingual clinical de-identification and synthetic PII generation.

\section*{Acknowledgments}

We thank Modal for providing the compute resources used to run our experiments.

\bibliographystyle{ws-procs11x85}
\bibliography{references}

\end{document}